\documentclass{article}

\usepackage{microtype}
\usepackage{graphicx}
\usepackage{subfigure}
\usepackage{subcaption}
\usepackage{booktabs} 
\usepackage{tablefootnote}

\usepackage{amsmath}
\usepackage{amssymb}
\usepackage{mathtools}
\usepackage{amsthm}

\theoremstyle{plain}

\theoremstyle{definition}

\theoremstyle{remark}

\usepackage[preprint]{neurips_2025}

\usepackage[utf8]{inputenc} 
\usepackage[T1]{fontenc}    
\usepackage{hyperref}       
\usepackage{url}            
\usepackage{booktabs}       
\usepackage{amsfonts}       
\usepackage{nicefrac}       
\usepackage{microtype}      
\usepackage{xcolor}         

\title{What Matters in LLM-generated Data: Diversity and Its Effect on Model Fine-Tuning}

\author{%
  Yuchang~Zhu\thanks{Equal Contribution.} \\
  Sun Yat-sen University\\
  \texttt{zhuych27@mail2.sysu.edu.cn} \\
  \And
  Huazhen~Zhong\footnotemark[1] \\
  Sun Yat-sen University \\
  \texttt{zhonghzh9@mail2.sysu.edu.cn} \\
  \And
  Qunshu~Lin\footnotemark[1] \\
  Abaka AI \\
  \texttt{linskysuka@gmail.com} \\
  \And
  Haotong~Wei \\
  Sun Yat-sen University \\
  \texttt{weiht5@mail2.sysu.edu.cn} \\
  \And
  Xiaolong~Sun \\
  Sun Yat-sen University \\
  \texttt{sunxlong@mail2.sysu.edu.cn} \\
  \And
  Zixuan~Yu \\
  Sun Yat-sen University \\
  \texttt{yuzx25@mail2.sysu.edu.cn} \\
  \And
  Minghao~Liu \\
  2077AI \\
  \texttt{dreamforever.liu@gmail.com} \\
  \And
  Zibin~Zheng \\
  Sun Yat-sen University \\
  \texttt{zhzibin@mail.sysu.edu.cn} \\
  \AND
  Liang~Chen\thanks{Corresponding Author.} \\
  Sun Yat-sen University \\
  \texttt{chenliang6@mail.sysu.edu.cn} \\ 
}

\begin{document}

\maketitle

\begin{abstract}
  With the remarkable generative capabilities of large language models (LLMs), using LLM-generated data to train downstream models has emerged as a promising approach to mitigate data scarcity in specific domains and reduce time-consuming annotations. However, recent studies have highlighted a critical issue: iterative training on self-generated data results in \textit{model collapse}, where model performance degrades over time. Despite extensive research on the implications of LLM-generated data, these works often neglect the importance of data diversity, a key factor in data quality. In this work, we aim to understand the implications of the diversity of LLM-generated data on downstream model performance. Specifically, we explore how varying levels of diversity in LLM-generated data affect downstream model performance. Additionally, we investigate the performance of models trained on data that mixes different proportions of LLM-generated data, which we refer to as synthetic data. Our experimental results show that, with minimal distribution shift, moderately diverse LLM-generated data can enhance model performance in scenarios with insufficient labeled data, whereas highly diverse generated data has a negative impact. 
We hope our empirical findings will offer valuable guidance for future studies on LLMs as data generators.
\end{abstract}

\section{Introduction}
\label{sec:intro}
The proliferation of large language models (LLMs)~\cite{achiam2023gpt,dubey2024llama} has increased innovation in employing them as data generators~\cite{wang2022self,yu2024large}. Inspired by these innovations, recent studies~\cite{ren2024learn,li2024api,wang2024codeclm} have leveraged this generated data, referred to as \textit{LLM-generated data}, to fine-tune downstream models, thereby tackling data scarcity issues in specific domains or reducing the need for time-consuming human annotations. Despite significant progress, some advancements~\cite{shumailov2024ai,shumailov2023curse} have revealed that models trained on LLM-generated data inevitably exhibit performance degradation, a phenomenon known as \textit{model collapse}.

With the development of LLMs, LLM-generated data is expected to become ubiquitous within the web ecosystem. Consequently, next-generation models will inevitably be trained on LLM-generated data or synthetic data that combines real-world data with LLM-generated data. In this regard, Shumailov et al.~\cite{shumailov2023curse} were the first to observe model collapse under the \textit{replace} scenario, where each successive generation model is trained on fully generated data from its predecessor, except for the first generation model, which is trained on real-world data. This phenomenon has raised increasing concerns about the impact of LLM-generated data, inspiring two primary research streams: one~\cite{gerstgrasser2024model,dohmatob2024strong,dohmatob2024tale} further investigating the impact of LLM-generated data under different scenarios and its theoretical foundations, and another~\cite{zhu2024synthesize,kuo2024not,feng2024beyond} focusing on fixing model collapse. 

On the one hand, existing studies empirically and theoretically verify whether model collapse occurs under various scenarios. For example, Gerstgrasser et al.~\cite{gerstgrasser2024model} demonstrate that model collapse can be avoided under the \textit{accumulate} scenario, where real-world data and LLM-generated data are accumulated to train next-generation models. Conversely, Dohmatob et al.~\cite{dohmatob2024strong} reveal that even incorporating a small proportion of LLM-generated data, such as 1\%, into training data can still lead to model collapse. Additionally, several studies empirically~\cite{seddik2024bad,bansal2024smaller} or theoretically~\cite{dohmatob2024tale,dohmatob2024model} investigate model collapse in various settings. In contrast, foundational LLMs, such as Llama 3~\cite{dubey2024llama}, incorporate LLM-generated data during their post-training data stage, thereby demonstrating the utility of such data. 
On the other hand, prior works~\cite{chen2024unveiling} focus on reducing distributional misalignment between LLM-generated data and real-world data. In this regard, existing studies leverage weighted-loss~\cite{kuo2024not}, unlearning~\cite{chen2024unveiling}, and token editing~\cite{zhu2024synthesize} to achieve this objective. Diverging from these studies, Feng et al.~\cite{feng2024beyond} propose verifying LLM-generated data to prevent model collapse, which follows the core idea that appropriate data selection results in optimal model performance.

Despite notable advancements, there is still no consensus on whether LLM-generated data causes model collapse and how to mitigate its effects. Dohmatob's work~\cite{dohmatob2024strong} highlights the necessity of maintaining the quality of LLM-generated data at the same level as real-world data. However, current studies~\cite{shumailov2024ai,gerstgrasser2024model} often neglect the consistency of data quality. As the core of data quality, the diversity of LLM-generated data, which presents sample richness and variation, has not been thoroughly explored in relation to model performance. Prior works~\cite{guo2023curious,briesch2023large} suggest that models trained on self-generated data suffer from a decline in output diversity. Thus, a natural question emerges: 
\begin{center}
\textit{How does the diversity of LLM-generated data impact next-generation models?}
\end{center}

In this work, we investigate the impact of LLM-generated data on trained models from a diversity perspective. We first explore the performance of models trained on LLM-generated data with varying levels of diversity, as well as synthetic data that mixes different proportions of LLM-generated data. Secondly, we further examine the correlation between model performance and different factors, such as model size, dataset size, model architecture, and model size. 
Furthermore, different from prior works that focus on the pre-training of LLMs, our work aims to study the impact of LLM-generated data's diversity in supervised fine-tuning (SFT), which are more closely aligned with practical applications of LLMs. To avoid the impact of other data quality, e.g., fluency and distribution, in our experiments, we ensure that these properties remain consistent with the real-world data. Our contributions and findings can be summarized as follows:
\begin{itemize}
    \item We explore model performance on LLM-generated data with varying diversity, akin to the replace scenario in~\cite{shumailov2024ai}. We find that while highly diverse LLM-generated data enhances performance, a distribution shift negatively impacts performance even with high diversity.
    \item We explore model performance on synthetic datasets with fixed and variable size, akin to the accumulate scenario in~\cite{gerstgrasser2024model}. The findings reveal that without a distribution shift, integrating moderately diverse data into fields with limited labeled data can boost performance.
    \item We further investigate scaling behaviors corresponding to varying dataset size, model architecture, and model size. The experimental results show consistent outcomes across three factors, indicating the generality of our findings.
\end{itemize}

\section{Related Work}
\label{sec:related_work}
Limited by space, we present the literature review of LLM-based data generation in Appendix~\ref{apd:related_work}.

\subsection{Impact of LLM-generated Data}
Shumailov et al.~\cite{shumailov2023curse,shumailov2024ai} reveal that AI models encounter model collapse when recursively training on LLM-generated data.
To explain model collapse, they offer a theoretical analysis indicating the absence of tails of the original distribution. Motivated by this, a series of follow-up studies theoretically~\cite{dohmatob2024tale,dohmatob2024strong} and empirically~\cite{bansal2024smaller,seddik2024bad} explore model collapse when using LLM-generated data. Dohmatob et al.~\cite{dohmatob2024tale} first analyze the change in scaling laws with the use of LLM-generated data and observe a wide range of decay phenomena. They further study this phenomenon in the setting of high-dimensional regression~\cite{dohmatob2024model}. Instead of recursively training on LLM-generated data, Gerstgrasser et al.~\cite{gerstgrasser2024model} theoretically and empirically demonstrate the avoidance of model collapse under an accumulate scenario, where next-generation models are trained on a combination of real-world data and LLM-generated data from predecessor models. Building on this, Dey et al.~\cite{dey2024universality} verify the universality of the accumulate scenario to avoid model collapse from a theoretical perspective, and Kazdan et al.~\cite{kazdan2024collapse} investigate model collapse between replace and accumulate scenarios. However, a recent study~\cite{dohmatob2024strong} indicates that model collapse cannot be avoided even if a small proportion of generated data, such as 1\%, is mixed into the training data. Some studies investigate the impact of LLM-generated data under a fixed inference budget~\cite{bansal2024smaller} or from a statistical perspective~\cite{seddik2024bad}. To avoid model collapse, recent studies leverage unlearning~\cite{chen2024unveiling}, token editing~\cite{zhu2024synthesize}, weighted-loss~\cite{kuo2024not}, and verification~\cite{feng2024beyond}. In summary, studies on the impact of LLM-generated data are still in their infancy and show significant differences in their scenarios, leading to distinctive conclusions.

Our study uniquely focuses on the theoretical and empirical analysis of model collapse from a diversity perspective, emphasizing the critical role of data quality. While related studies exist, they differ significantly in their focus. Specifically, Chen et al.~\cite{chen2024diversity} propose a LLM-based diversity evaluation method, aiming to verify its effectiveness. Conversely, our work examines the impact of LLM-generated data diversity in SFT scenarios. Guo et al.~\cite{guo2023curious} and Briesch et al.~\cite{briesch2023large} empirically demonstrate a decrease in output diversity for models trained on LLM-generated data. In contrast to these studies, our work aims to the diversity of the input data used for model training and evaluates the performance of the model on downstream tasks, rather than the diversity of its output.

\section{Preliminaries}
\label{sec:prelimi}
\subsection{Notations}
\label{subsec:notations}
To enhance the clarity of our paper, we present notations related to this paper in this subsection. Let $\mathcal{D}=\{(x_{i}, y_{i})\}_{i=1}^{n}$ denote a real-world dataset consisting of $n$ input-output pairs $(x_{i}, y_{i})$, where $x_{i}$ and $y_{i}$ represent the input and output texts, respectively. Notably, $x_{i}$ and $y_{i}$ may vary depending on the task associated with $\mathcal{D}$. We also define a LLM-generated dataset and a synthetic dataset as $\overline{\mathcal{D}}=\{(\overline{x}_{i}, \overline{y}_{i})\}_{i=1}^{n}$, $\Tilde{\mathcal{D}}=\{(\Tilde{x}_{i}, \Tilde{y}_{i})\}_{i=1}^{n}$, respectively. The LLM-generated dataset consists solely of generated data, while the synthetic dataset is a combination of LLM-generated data and real-world data. Given a LLM model $\mathcal{\mathcal{M}}$ serving as a dataset generator, the process of generating or augmenting a dataset can be formulated as $\overline{\mathcal{D}} \leftarrow \mathcal{M}(\mathcal{T})$. Here, $\mathcal{T}=\operatorname{Task}(t,s,s^{'})$ is the input text that prompts $\mathcal{M}$ to generate the desired data. The function $\operatorname{Task}(\cdot)$ preprocesses the dataset task $t$, the seed example $s$, and the supplementary materials $s^{'}$ to create this input text, with $s^{'}$ being optional and serving as a constraint for generation. 


\subsection{Data Generation and Diversity Evaluation}
\label{subsec:data_gene}

\subsubsection{Diversity-controlling Data Generation}
\label{subsubsec:data_gen}
\begin{figure}[!tbp]
    \centering
    \includegraphics[width=0.95\linewidth]{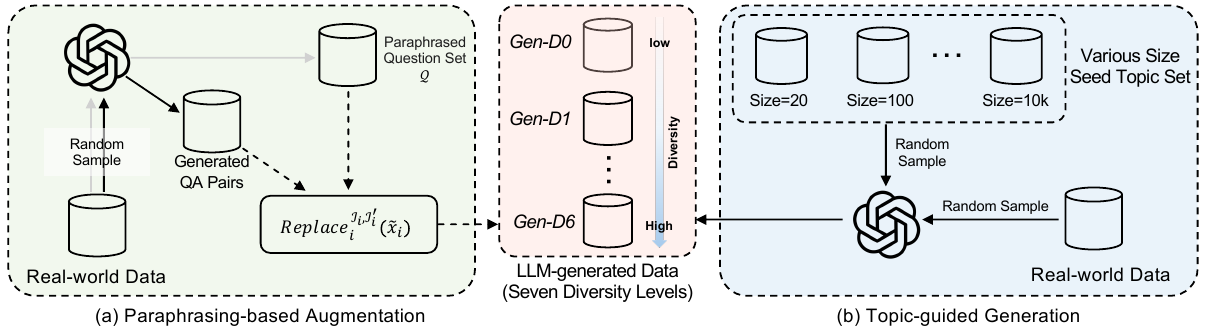}
    \caption{Diversity-controlling data generation pipelines in this study. (a) Paraphrasing-based augmentation involves repeatedly paraphrasing a sample and replacing some of the generated samples with a corresponding number of paraphrased samples. (b) Topic-guided generation constructs datasets with different levels of diversity by varying the number of candidate seed topics.}
    \label{fig:generate}
\end{figure}

To investigate the impact of LLM-generated data from a diversity perspective, we need to control the diversity of LLM-generated data. While previous studies~\cite{caccia2018language,tevet2020evaluating} have shown a positive correlation between softmax temperature and generated text, this relationship is limited to scenarios with multiple generations from the same context. To avoid redundancy, we perform data generation once for each context. Figure~\ref{fig:generate} illustrates our diversity-controlling data generation pipelines, which include paraphrasing-based augmentation and topic-guided generation. These pipelines enable us to construct LLM-generated data with different diversity levels. 

\textbf{Paraphrasing-based Augmentation.} Some tasks rely on prerequisites and are unsuitable for direct sample creation using LLMs. For example, the long context question-answer (QA) task requires answering questions based on a given lengthy context, making it impossible to create new QA pairs without this context. Thus, we utilize paraphrasing-based augmentation to achieve diversity control for the data construction of these tasks. The core idea behind this approach is to control the proportion of similar samples within LLM-generated datasets. In particular, paraphrasing-based augmentation involves repeatedly paraphrasing a sample and replacing some of the generated samples with a corresponding number of paraphrased (similar) samples that maintain similar semantics. 

Take the long context QA task as an example. Given a long context QA dataset $\mathcal{D}=\{(x_{i}, y_{i})\}_{i=1}^{n}$, where $x_{i}=\{[l_{i},q_{ik}]\}_{k=1}^{K}$ represents the set of the concatenation of the long context $l_{i}$ and its questions $q_{ik}$, and $y_{i}=\{a_{ik}\}_{k=1}^{K}$ represents the corresponding set of answers. As shown in Figure~\ref{fig:generate} (a), for each long context $l_{i}$, the paraphrasing-based augmentation first prompts LLMs to generate new QA pairs, which takes $l_{i}$ as $s^{'}$ and randomly samples real-world QA pairs as $s$. Here, the detailed prompt is provided in Figure~\ref{fig:prompt_new_qa} of Appendix~\ref{subsec:prompt_paraphrase}. For all $l_{i}$, we construct a synthetic dataset $\Tilde{\mathcal{D}}=\{(\Tilde{x}_{i}, \Tilde{y}_{i})\}_{i=1}^{n}$ with generated QA pairs, where $\Tilde{x}_{i}=\{[l_{i},\Tilde{q}_{ik}]\}_{k=1}^{K}$ and $\Tilde{y}_{i}=\{\Tilde{a}_{ik}\}_{k=1}^{K}$. Then, to achieve datasets with different diversity levels, the paraphrasing-based augmentation further paraphrases each question $q_{ik}$ by using a QA pair $(q_{ik},a_{ik})$ as $s^{'}$ and randomly sampling a QA pair $(q_{jk},a_{jk}),j\neq i$ as $s$. This process yields a paraphrased question subset $\mathcal{P}_{ik}=\{\overline{q}_{ik}^{1},\overline{q}_{ik}^{2},...,\overline{q}_{ik}^{m_{p}}\}$ for $q_{ik}$ and $a_{ik}$, where $m_{p}$ is the number of paraphrased questions. For all $l_{i}$, we can construct a paraphrased question set $\mathcal{Q}=\{\mathcal{P}_{i1},...,\mathcal{P}_{iK}\}_{i=1}^{n}$, which can be formulated as follows:
\begin{equation}
\label{eq:paraphrased_q}
    \mathcal{Q}\leftarrow \mathcal{M}(\operatorname{Task}(t,(q_{ik},a_{ik}),(q_{jk},a_{jk}))), i, j=1,2,...,n, \text{and}\ j\neq i,
\end{equation}
where the detailed prompt of the above process is presented in Figure~\ref{fig:prompt_para_q} of Appendix~\ref{subsec:prompt_paraphrase}.

Based on $\mathcal{Q}$, we randomly replace some generated QA pairs in $\Tilde{\mathcal{D}}$ with paraphrased ones to control the diversity of the final synthetic dataset. Take the replacement of questions as an example, let $m_{r}=m_{p}\times r$ denote the number of replaced generated questions in $\Tilde{\mathcal{D}}$, with $\mathcal{I}_{i}\subseteq \{1,2,..,K\}$ and $\mathcal{I}_{i}^{'}\subseteq \{1,2,..,K\}$ representing the index sets of questions to be replaced and those used for replacement, respectively. Here, $|\mathcal{I}_{i}^{'}|=r, r\in \mathbb{Z}$ and $|\mathcal{I}_{i}|=m_{r}$. For each $x_{i}$, the replacement operation can be formulated as follows:
\begin{equation}
\label{eq:replacement1}
    \operatorname{Replace}_{i}^{\mathcal{I}_{i},\mathcal{I}_{i}^{'}}(\Tilde{x}_{i})=\{[l_{i},\hat{q}_{ik}]\}_{k=1}^{K},
\end{equation}
where 
\begin{equation}
\label{eq:replacement2}
    \hat{q}_{ik} = 
    \begin{cases}
    \Tilde{q}_{ik}, & k \notin \mathcal{I}_i \cup \mathcal{I}_i^{'} \\
    \mathcal{P}_{ik}, & k \in \mathcal{I}_i^{'} \\
    \text{removed (replaced)}, & k \in \mathcal{I}_i
    \end{cases}
\end{equation}

Additionally, the replacement of the answer aligns with the aforementioned question replacement. Based on Equations~\eqref{eq:replacement1} and~\eqref{eq:replacement2}, we can vary $m_{r}$ to construct synthetic data with different degrees of diversity. Here, a larger $m_{r}$ diminishes the diversity of the synthetic data. In our experiments, we employ paraphrasing-based augmentation in the construction of synthetic datasets for the long context QA task, with NarrativeQA~\cite{narrativeqa} serving as $\mathcal{D}$.

\textbf{Topic-guided Generation.} Prior works~\cite{chen2024diversity} have shown a positive relationship between the number of seed topics and the diversity of generated data. Building on this core idea, topic-guided generation controls the diversity of generated datasets by varying the size of the seed topic set. Specifically, we first prompt LLMs to generate a set of $m_{t}$ topics $\mathcal{A}=\{\mathcal{S}_{1}, \mathcal{S}_{2}, ..., \mathcal{S}_{m_{t}}\}$. The detailed prompt of topic generation is presented in Appendix~\ref{subsec:prompt_seed_topics}. Given a real-world dataset $\mathcal{D}=\{(x_{i}, y_{i})\}_{i=1}^{n}$, for each generation, a seed topic $\mathcal{S}_{j}$ is randomly sampled from $\mathcal{A}$, along with several seed samples $\mathcal{D}_{s}$ from $\mathcal{D}$. We then construct the input text (also known as the prompt) for the dataset generator $\mathcal{M}$, denoted by $\mathcal{T}=\operatorname{Task}(t,\mathcal{D}_{s},\mathcal{S}_{j})$, and prompt $\mathcal{M}$ to generate the LLM-generated dataset $\overline{\mathcal{D}}$. To maintain distribution consistency between $\overline{\mathcal{D}}$ and $\mathcal{D}$, we ensure that the token count of the generated data matches the seed samples through the generation prompt. As shown in Figure~\ref{fig:generate} (b), different diversity levels of LLM-generated data can be achieved based on varying sizes of seed topic sets. Additionally, we apply topic-guided generation for story completion data generation, using ROCStories~\cite{rocstories} as $\mathcal{D}$. Limited by space, the detailed prompt of data generation is provided in Appendix~\ref{subsec:prompt_topic}.

\subsubsection{Diversity Evaluation}
\label{subsubsec:diver_eval}            
To assess the diversity of synthetic data, we utilize the \textit{Distinct-n} metric, which quantifies diversity by calculating the proportion of unique n-grams (e.g., 2-grams and 3-grams) within the generated text~\cite{distinctn}. This metric computes the ratio of distinct n-grams to the total number of n-grams. Formally, the evaluation of distinct-n for $\mathcal{D}$ can be formulated as follows:
\begin{equation}
    \label{eq:distinct_n}
    \operatorname{Distinct-n}(\mathcal{D}) = \frac{|\operatorname{Unique}(\operatorname{n-grams}(\operatorname{Concat}(\mathcal{D})))|}{|\operatorname{n-grams}(\operatorname{Concat}(\mathcal{D}))|},
\end{equation}
where $\operatorname{n-grams}$, $\operatorname{Unique}$, and $\operatorname{Concat}$ represent the n-grams, de-overlap process, and concatenate operation, respectively. Notably, for the long context QA dataset, we first calculate distinct-n for each long context and then calculate the sum of these distinct-n values across all long contexts within the dataset. Specifically, given a long context QA dataset $\mathcal{D}=\{(x_{i}, y_{i})\}_{i=1}^{n}$, where $x_{i}=\{[l_{i},q_{ik}]\}_{k=1}^{K}$ and $y_{i}=\{a_{ik}\}_{k=1}^{K}$, the diversity of $\mathcal{D}$ can be calculated as follows:
\begin{equation}
    \label{eq:divers_qa}
    Diversity=\sum_{i=1}^{n}{\operatorname{Distinct-n}(\{(q_{iK},a_{iK})\}_{k=1}^{K})}.
\end{equation}

For the story completion dataset $\mathcal{D}=\{(x_{i}, y_{i})\}_{i=1}^{n}$, the calculation can be formulated as follows:
\begin{equation}
    \label{eq:divers_story}
    Diversity=\operatorname{Distinct-n}(\{(x_{i}, y_{i})\}_{i=1}^{n}).
\end{equation}



\section{Experiments}
\label{sec:exp}
In this section, we first investigate the effects of LLM-generated data across different diversity levels and mixing ratios. We then explore scaling behaviors corresponding to varying dataset sizes, model architectures, and model sizes. Due to space constraints, experiments involving different model architectures and sizes are detailed in Appendix~\ref{apd:add_exp}.

\subsection{Experimental Settings}
\label{subsec:exp_set}
Limited by space, we present details of datasets in Appendix~\ref{sec:datasets}.

\subsubsection{Evaluation Settings}
\label{subsubsec:eval_settings}
In line with previous studies~\cite{gerstgrasser2024model}, we use test loss as our performance metric to ensure consistency across various downstream tasks. For diversity evaluation, we leverage \textit{Distinct-5} in Equations~\eqref{eq:divers_qa} and~\eqref{eq:divers_story} for the long context QA and story completion datasets, respectively. We normalize the diversity evaluation results within a comparison group by dividing by the maximum value within the group. Due to variations in dataset size, we did not normalize the diversity results in Figures~\ref{fig:mix_vary_qa} and~\ref{fig:mix_vary_story}.

\subsubsection{Implementation Details}
\label{subsubsec:imple_details}
We train GPT-2-124M, Llama 3.2-1B, and Llama 3.1-8B on question-answering and story completion tasks. For a fair comparison, we utilize the same hyperparameter for all model fine-tuning. Specifically, we employ LoRA with a rank of 8 and an alpha parameter of 16 for scaling. The learning rate, weight decay, and number of epochs are set to 1e-4, 1e-4, and 3, respectively. Additionally, we save the model at the final training step to evaluate the loss on the test set of real-world datasets. For diversity evaluation of training datasets, we utilize the distinct-n metric as detailed in Section~\ref{subsubsec:diver_eval}. All experiments are conducted on 8$\times$ NVIDIA RTX 4090 GPU with 24 GB memory.

\subsection{Model Performance on LLM-generated Data with Various Diversity}
\label{subsec:var_diver}
\textbf{Setup.} We investigate the impact of LLM-generated data with varying diversity levels on model performance. As shown in Appendix~\ref{sec:datasets}, we construct seven diversity levels of LLM-generated data, named from \textit{Gen-D0} to \textit{Gen-D6}, representing the lowest to highest diversity, respectively. We ensure that the volume of LLM-generated dataset samples matches that of real-world datasets. The datasets for question-answering and story completion tasks contain 32,291 and 30,000 samples, respectively. We fine-tune three LLMs of different sizes: GPT-2-124M, Llama 3.2-1B, and Llama 3.1-8B, referred to as 124M, 1B, and 8B. Then, we evaluate our trained models on the test set of the real-world dataset.

\begin{table}[t]
\caption{Dataset statistic of NarrativeQA and ROCStories.}
\label{tab:data_stat}
\begin{center}
\begin{small}
\begin{sc}
\begin{tabular}{lcc}
\toprule
Datasets & Train Set (Num.) & Test Set (Num.) \\
\midrule
NarrativeQA\tablefootnote{\url{https://huggingface.co/datasets/deepmind/narrativeqa}}    & 32747& 10557 \\
ROCStories\tablefootnote{\url{https://huggingface.co/datasets/mintujupally/ROCStories}} & 78528 & 19633 \\
\bottomrule
\end{tabular}
\end{sc}
\end{small}
\end{center}
\end{table}

\begin{figure}[!tbp]
    \centering
    \begin{subfigure}
        \centering
        \includegraphics[width=0.38\linewidth]{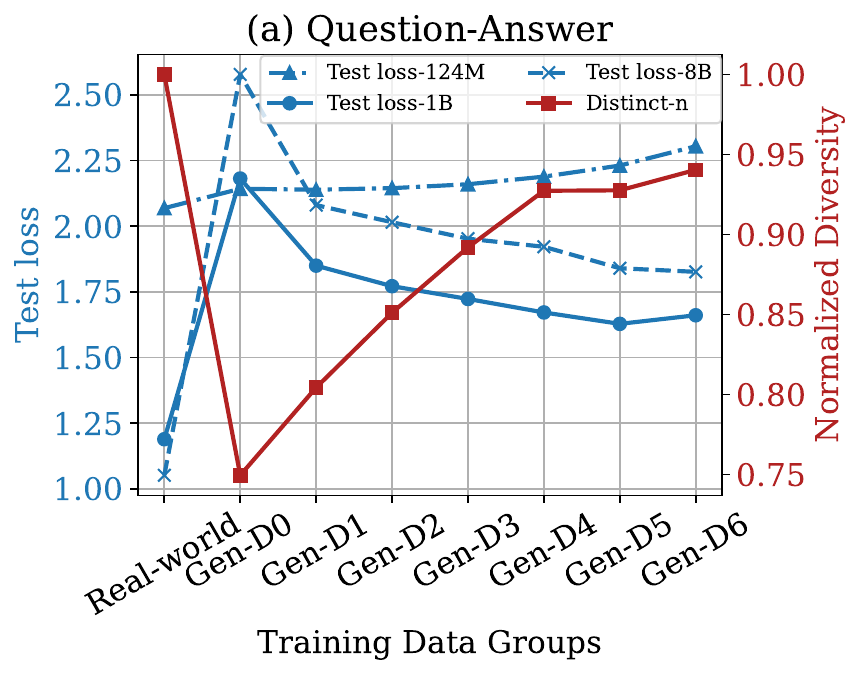}
        \label{fig:divers_qa}
    \end{subfigure}
    \begin{subfigure}
        \centering
        \includegraphics[width=0.38\linewidth]{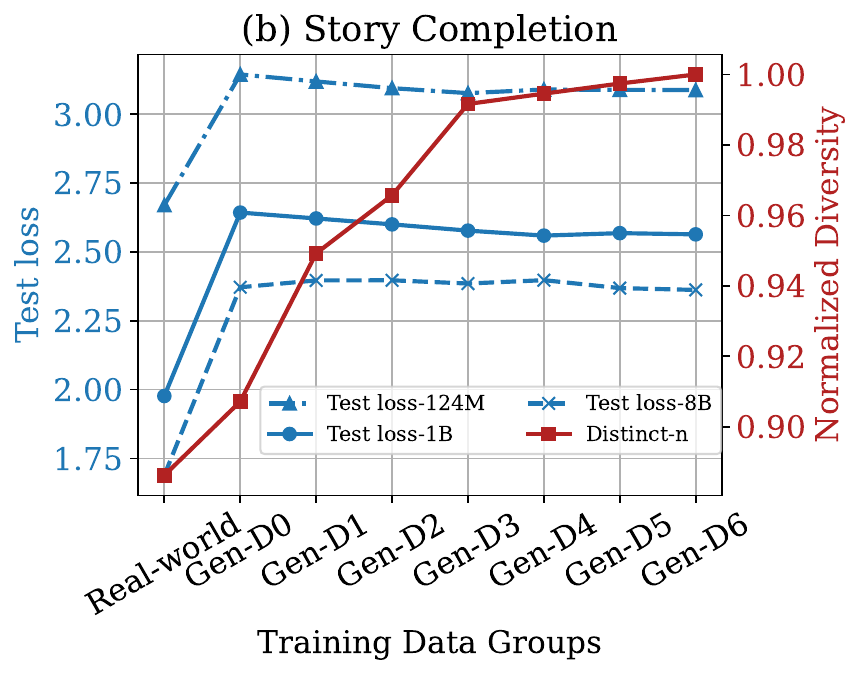}
        \label{fig:divers_story}
    \end{subfigure}
    \caption{Correlation between model performance and the diversity of LLM-generated data. (a), (b) Model test loss and normalized diversity on the question-answering task (story completion task).}
    \label{fig:all_utility_divers}
\end{figure}

\begin{figure}[!tb]
    \centering
    \begin{minipage}[b]{0.52\textwidth}
        \centering
        \includegraphics[width=\linewidth]{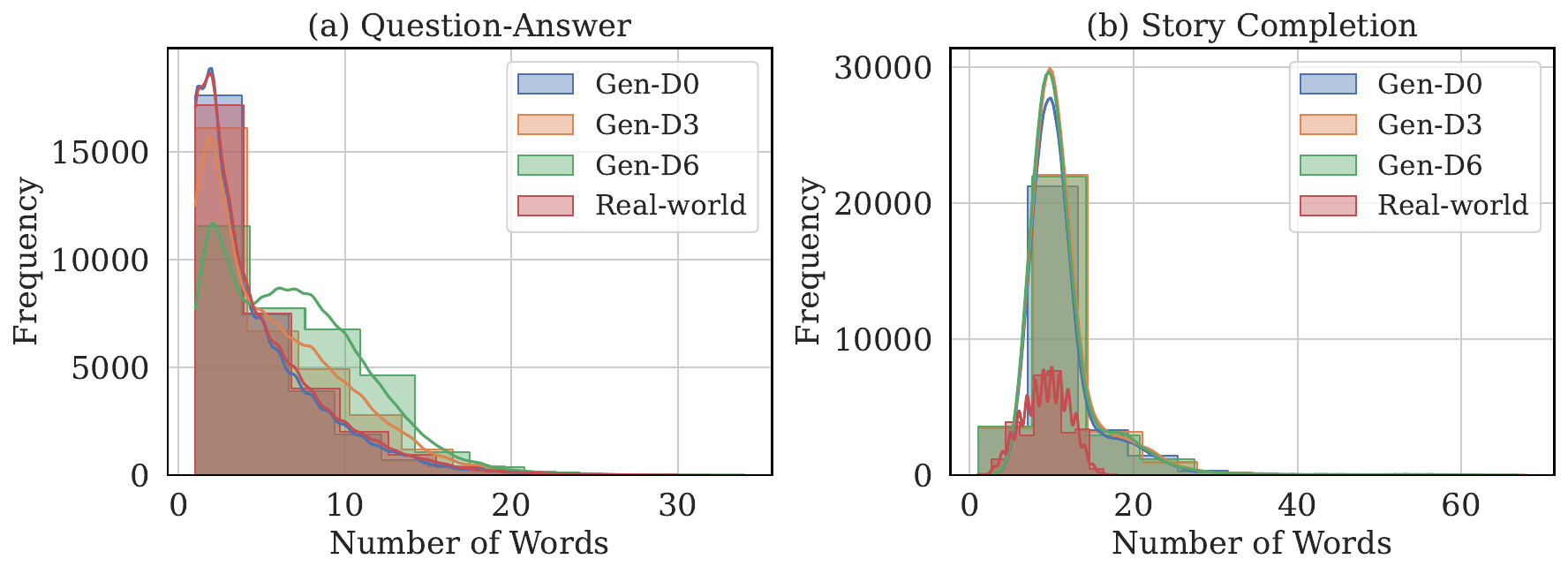}
        \caption{Comparison of answer length distribution between real-world and LLM-generated datasets.}
        \label{fig:len_ana}
    \end{minipage}
    \hfill
    \begin{minipage}[b]{0.46\textwidth}
        \centering
        \includegraphics[width=\linewidth]{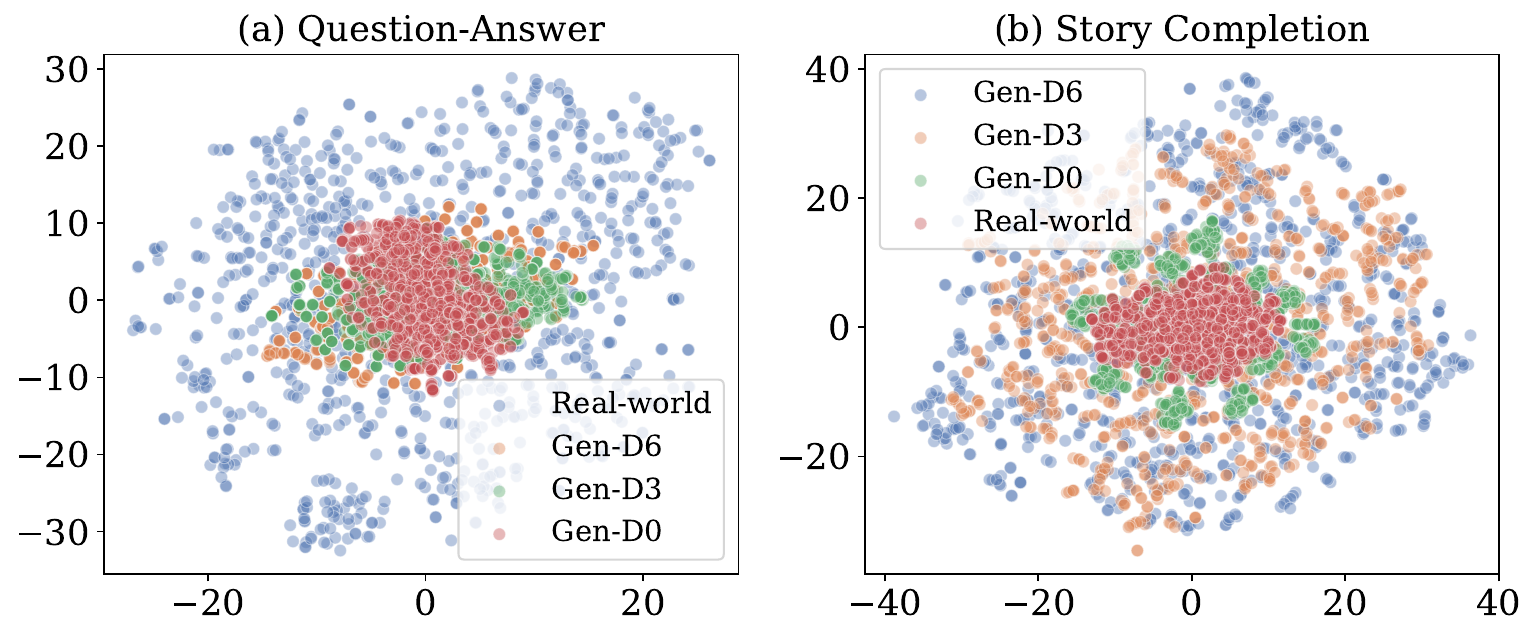}
        \caption{T-SNE visualization of embeddings for real-world and LLM-generated datasets. 
        }
        \label{fig:embed_ana}
    \end{minipage}
\end{figure}

\textbf{Results.} The model test loss and normalized diversity are illustrated in Figure~\ref{fig:all_utility_divers}. Our experimental results for both question-answering and story completion tasks reveal a similar trend: increasing the diversity of LLM-generated datasets leads to improved model performance, as evidenced by a decrease in model test loss. However, \textbf{regardless of how the diversity of LLM-generated data compares to real-world datasets, models trained on LLM-generated data are consistently inferior to those trained on real-world datasets}. In the question-answering task, the paraphrasing-based augmentation results in lower diversity in LLM-generated data compared to real-world data, limiting the generalization capabilities of the trained models. This prompts the question: \textit{If the diversity of LLM-generated data exceeds that of real data, can better performance be achieved?} Figure~\ref{fig:all_utility_divers} (b) offers insight into this, indicating that even with higher diversity in LLM-generated datasets, the trained model still underperforms compared to one trained on the real-world dataset. 

To gain deeper insights into our observations, we compare the distributions of LLM-generated and real-world datasets. Specifically, we begin by analyzing the answer lengths in both real-world and LLM-generated datasets. We only present the answer lengths of the real-world, Gen-D0, Gen-D3, and Gen-D6 datasets, as the other datasets show similar results. As shown in Figure~\ref{fig:len_ana}, the answer lengths for both real-world and LLM-generated datasets remain consistent in the question-answering task. However, a notable difference in answer length is observed between real-world and LLM-generated datasets in the story completion task. This phenomenon suggests a distribution shift in LLM-generated datasets, which may be a potential reason why the diversity of generated data is greater than that of real data, yet the trained model performs worse than one trained on real data. In the question-answering task, we find that more diverse datasets tend to have longer answer lengths.

\begin{table}[!t]
\caption{Distribution distance (KL Divergence) between real-world and LLM-generated datasets.}
\label{tab:distribution_dist}
\begin{center}
\begin{small}
\begin{sc}
\begin{tabular}{lcc}
\toprule
Diversity & question-answering & Story Completion \\
\midrule
Gen-D0   & 0.0057  & 2.4507  \\
Gen-D1   & 0.0078  & 2.5120  \\
Gen-D2   & 0.0089  & 2.5651  \\
Gen-D3   & 0.0088  & 2.7896  \\
Gen-D4   & 0.0111  & 2.7827  \\
Gen-D5   & 0.0095  & 2.7635  \\
Gen-D6   & 0.0099  & 3.0178  \\
\bottomrule
\end{tabular}
\end{sc}
\end{small}
\end{center}
\vskip -0.1in
\end{table}

To further explain the potential reasons behind Figure~\ref{fig:all_utility_divers} (b), we visualize sample embedding of real-world and LLM-generated datasets. Specifically, we concatenate question (story context) and answer (story completion) to show the distribution of datasets. Importantly, to avoid the impact of long contexts, we exclude the long context in the question-answering task during visualization. To clarify visualization, we randomly sample 1000 data points and obtain embeddings using \textsf{all-mpnet-base-v2}\tablefootnote{\url{https://huggingface.co/sentence-transformers/all-mpnet-base-v2}}. As shown in Figure~\ref{fig:embed_ana}, the embedding visualization reveals similar results to the diversity observed in Figure~\ref{fig:all_utility_divers}. Datasets with higher diversity correspond to larger embedding clusters in Figure~\ref{fig:embed_ana}. In the question-answering task, even the most diverse dataset, such as Gen-D6, covers only a small proportion of the real-world distribution, thereby limiting the performance of models trained on LLM-generated datasets. In the story completion task, we observe an opposite phenomenon compared to the question-answering task. LLM-generated datasets exhibit larger embedding clusters than real-world datasets while incorporating the embedding cluster of real-world data. Models trained on LLM-generated datasets should theoretically outperform those trained on real-world datasets. However, Figure~\ref{fig:all_utility_divers} (b) indicates that models trained on LLM-generated datasets exhibit inferior performance. To clarify this, we calculate the distribution distance between real-world and LLM-generated datasets, as shown in Table~\ref{tab:distribution_dist}. We find that LLM-generated data for the story completion task has a significant distribution distance from real-world distributions. This could be a key reason why higher dataset diversity does not result in better model performance. Thus, \textbf{while a highly diverse LLM-generated dataset may enhance model performance, a distribution shift can negatively impact performance even with high diversity.}

\subsection{Model Performance on Synthetic Data with Various Mixing Ratios}
\label{subsec:mix_ratio}

\textbf{Setup.} We investigate model performance on synthetic data with various mixing ratios under both fixed and variable dataset sizes. For fixed dataset sizes, we fine-tune three models of different sizes on synthetic datasets created by mixing real-world and LLM-generated data. We then evaluate these models using the same test set as in Section~\ref{subsec:var_diver}. We randomly sample 22,040 and 20,000 instances from the original real-world datasets for the question-answering and story completion tasks, respectively, using these subsets as the real-world data source. For the LLM-generated data source, we utilize the \textit{Gen-D6} dataset from Section~\ref{subsec:var_diver}. We create synthetic datasets with different mixing ratios by random sampling from real-world and LLM-generated data sources, ensuring the synthetic datasets match the sample size of the real-world subsets. For variable dataset sizes, we fix the real-world data to 2,204 and 2,000 samples for the question-answering and story completion tasks, respectively. Additionally, we fine-tune 124M, 1B, and 8B models on synthetic datasets of different sizes. Here, the mixing ratio $\rho$ is formulated as follows:
\begin{equation}
    \label{eq:mixing_ratio}
    \rho = \frac{|\mathcal{D}_{gen}|}{|\mathcal{D}_{real}|+|\mathcal{D}_{gen}|}\times 100\%,
\end{equation}
where $\mathcal{D}_{gen}$ and $\mathcal{D}_{real}$ represent data from LLM-generated and real-world subsets, respectively.

\begin{figure}[!tbp]
    \centering
    \begin{subfigure}
        \centering
        \includegraphics[width=0.38\linewidth]{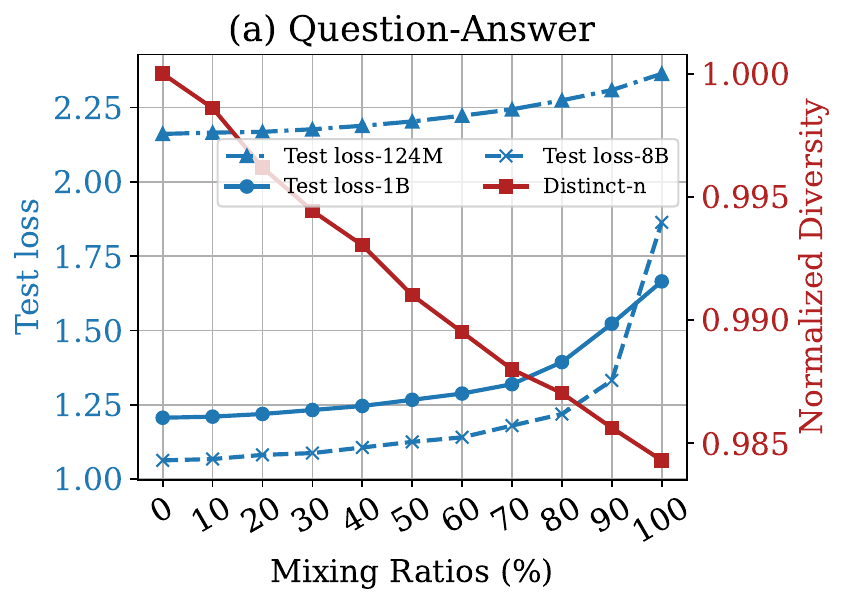}
        \label{fig:mix_qa}
    \end{subfigure}
    \begin{subfigure}
        \centering
        \includegraphics[width=0.38\linewidth]{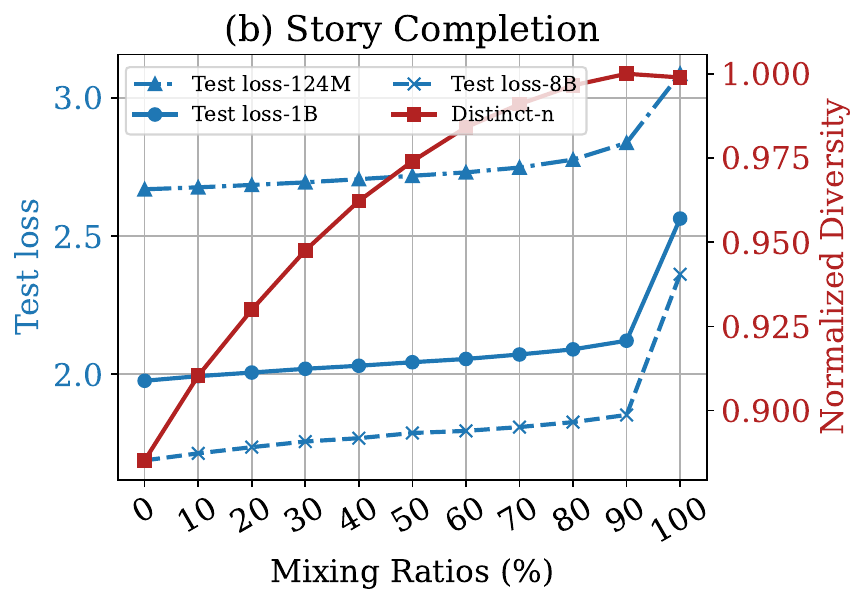}
        \label{fig:mix_story}
    \end{subfigure}
    \caption{Model performance and dataset diversity on synthetic datasets with fixing size. (a), (b) Experimental results on the question-answering task (story completion task).}
    \label{fig:mix_ratio_all}
\end{figure}

\textbf{Results.} As shown in Figure~\ref{fig:mix_ratio_all}, our results consistently show across two downstream tasks and three different models: \textbf{Regardless of whether the inclusion of LLM-generated data increases or decreases diversity, as the mixing ratio increases, i.e., the proportion of LLM-generated data rises, the model's performance gradually declines.} Even a small proportion of LLM-generated data, such as 10\%, incorporated into the training data leads to an increase in model test loss, consistent with previous studies~\cite{dohmatob2024strong}. The performance degradation in the question-answering task may be attributed to the limited patterns resulting from decreased diversity in synthetic datasets. In contrast, the performance drop in the story completion task might be linked to the distribution shift of LLM-generated data, as indicated in Table~\ref{tab:distribution_dist}. As the proportion of LLM-generated data rises, the distribution shift between synthetic and real-world datasets gradually increases.


\begin{figure}[!tb]
    \centering
    \includegraphics[width=0.9\linewidth]{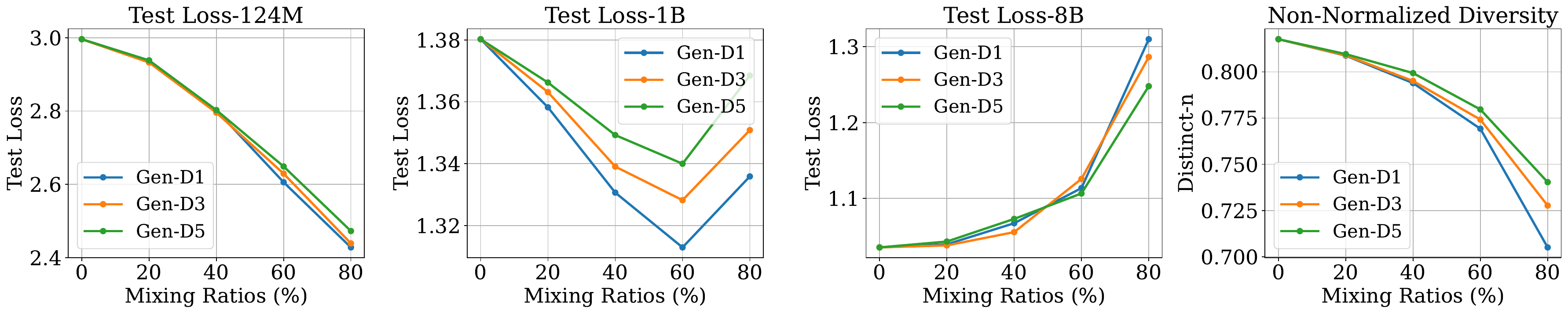}
    \caption{Model performance and dataset diversity on variable-size synthetic datasets for the QA task.}
    \label{fig:mix_vary_qa}
\end{figure}

\begin{figure}[!tb]
    \centering
    \includegraphics[width=0.9\linewidth]{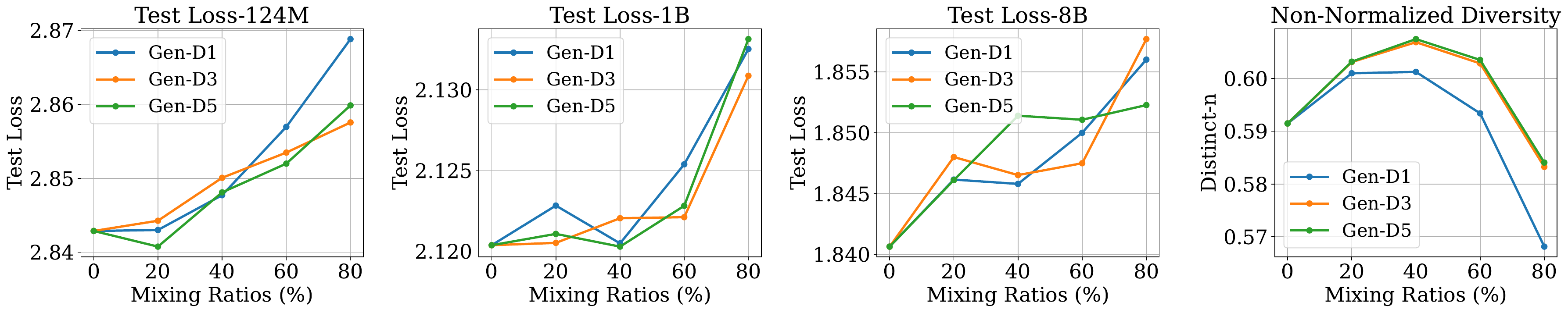}
    \caption{Model performance and dataset diversity on variable-size synthetic datasets for the story completion task.}
    \label{fig:mix_vary_story}
\end{figure}

We further investigate model performance on variable-size synthetic datasets by starting with a small amount of labeled real data and continuously adding LLM-generated data. Specifically, we employ multi-level diversity LLM-generated data used in Section~\ref{subsec:var_diver} as the source for data expansion. Due to the similar experimental results, we only present results where Gen-D1, Gen-D3, and Gen-D5 are used as expansion sources, as shown in Figure~\ref{fig:mix_vary_qa} and~\ref{fig:mix_vary_story}. Notably, we present non-normalized diversity results for varying sizes of training data. For the question-answering task, we find that as more generated data is added, the model's performance improves. However, when the proportion of generated data is high, such as 80\%, the performance of the 1B model declines. Additionally, lower-level diversity LLM-generated data is more effective in improving model performance, particularly for the 1B model. For the story completion task, we observe that the diversity of synthetic data initially increases and then decreases, indicating a significant distribution shift between LLM-generated and real data. Adding a small amount of generated data quickly enhances data diversity, but adding more leads to overlap with previously added data, reducing diversity. Furthermore, as the proportion of LLM-generated data increases, the model's performance gradually degrades, even though the inclusion of LLM-generated data enhances data diversity. This phenomenon is still related to the significant distribution shift between the LLM-generated data and the real data.

In summary, \textbf{ensuring no distribution shift, incorporating moderately diverse data to fields with insufficient labeled data can improve the model's performance}.

\subsection{Impact of Dataset Size}
\label{subsec:impact_data_size}

\begin{figure*}[!tb]
    \centering
    \includegraphics[width=\textwidth]{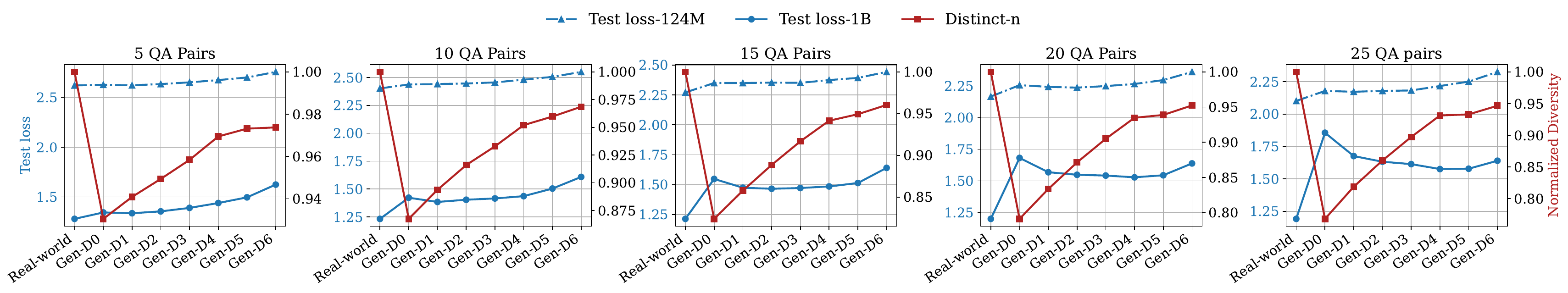}
    \caption{Model performance on QA datasets (real-world or LLM-generated datasets) of varying sizes. We vary the number of question-answering pairs to \{5, 10, 15, 20, 25\} for each long context.}
    \label{fig:data_size_qa}
\end{figure*}

\begin{figure*}[!tb]
    \centering
    \includegraphics[width=\textwidth]{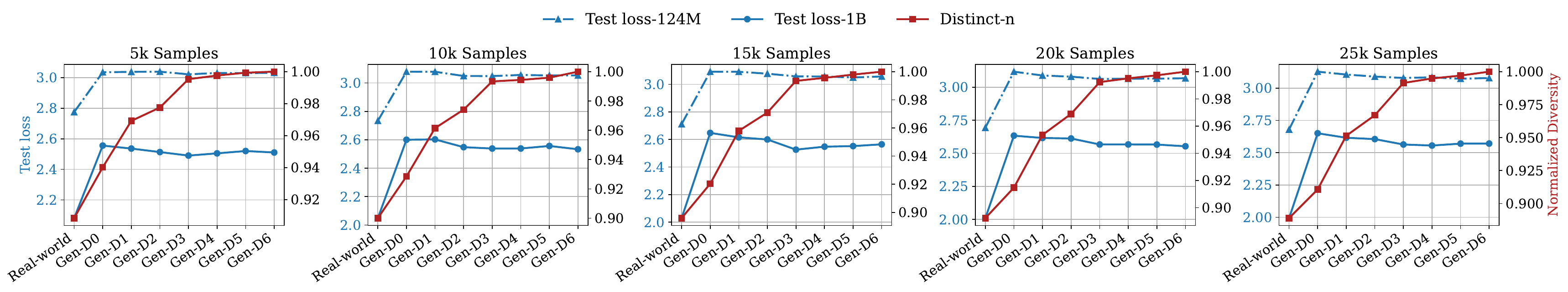}
    \caption{Model performance on story completion datasets (real-world or LLM-generated datasets) of varying sizes. We vary sample number of training datasets to \{5k, 10k, 15k, 20k, 25k\}.}
    \label{fig:data_size_story}
\end{figure*}

\textbf{Setup.} We generalize the experiments from Section~\ref{subsec:var_diver} to include various dataset sizes. For the QA task, we set the number of QA pairs to \{5, 10, 15, 20, 25\} for each long context. For the story completion task, we set the number of samples to \{5k, 10k, 15k, 20k, 25k\}. These different dataset sizes are randomly sampled from the training datasets in Section~\ref{subsec:var_diver}. To maintain computational efficiency and be environmentally friendly, we fine-tune only small models, including 124M and 1B models. We evaluate the trained models on the same datasets as in Section~\ref{subsec:var_diver}.

\textbf{Results.} As shown in Figures~\ref{fig:data_size_qa} and~\ref{fig:data_size_story}, we observe similar results to those shown in Figure~\ref{fig:all_utility_divers}. Specifically, for the QA task, 124M models perform worse as the diversity of LLM-generated datasets increases. Small models struggle with long context understanding, and more diverse training data exacerbates the fitting difficulty. Hence, 124M models show distinct performance patterns in contrast to 1B models. Conversely, the 1B model's performance initially improves but then declines with increased diversity of LLM-generated data. Meanwhile, we find that larger datasets lead to poorer performance for models trained on LLM-generated data, indicating that more LLM-generated data can be harmful. For the story completion task, we observe consistent performance trends across two models and five dataset sizes, aligning with the results in Figure~\ref{fig:all_utility_divers} (b). To sum up, we reproduce results shown in Figure~\ref{fig:all_utility_divers} across different dataset sizes, eliminating the impact of dataset size.

\section{Conclusion}
\label{sec:conclu}
In this paper, we investigate the implications of the diversity of LLM-generated data on downstream model performance. Specifically, we first introduce two data generation strategies to control the diversity of LLM-generated data. Using these strategies, we explore the model performance on LLM-generated datasets with varying levels of diversity and investigate the implications of synthetic datasets by combining real-world and LLM-generated data. To mitigate the impact of dataset size, we repeat the aforementioned experiments with various training dataset sizes. 

\textit{What is the impact of the diversity of generated data on the next-generation of models?} In Section~\ref{subsec:var_diver}, our observations reveal that while a highly diverse LLM-generated dataset may enhance model performance, a distribution shift negatively impacts performance even with high diversity. In Section~\ref{subsec:mix_ratio}, we observe that without a distribution shift, integrating moderately diverse data into fields with limited labeled data can boost model performance. In Section~\ref{subsec:impact_data_size}, we confirm that similar experimental results can be observed in different settings, indicating the versatility of our findings. Although our experimental results are effective, there is a limitation in our study. Ensuring that the data distribution is consistent between real-world and LLM-generated data is crucial due to the impact of the distribution shift on model performance.

\bibliography{refs}
\bibliographystyle{plain}

\newpage
\appendix

\section{Details of LLM-generated Data}
\label{apd:prompt}

\subsection{Prompt for Paraphrasing-based Augmentation.}
\label{subsec:prompt_paraphrase}

\begin{figure}[!ht]
    \centering
    \includegraphics[width=0.95\textwidth]{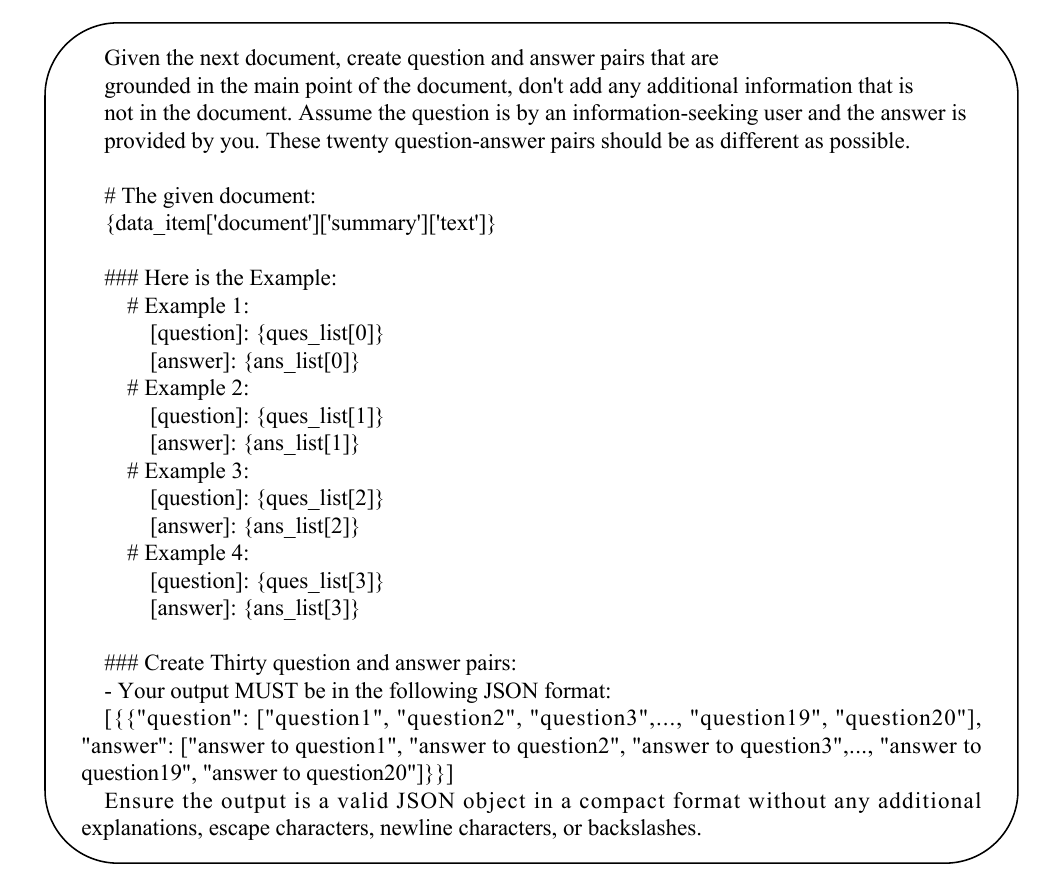}  
    \caption{Prompt for new QA pairs generation.}
    \label{fig:prompt_new_qa}
\end{figure}

\begin{figure}[!ht]
    \centering
    \includegraphics[width=0.95\textwidth]{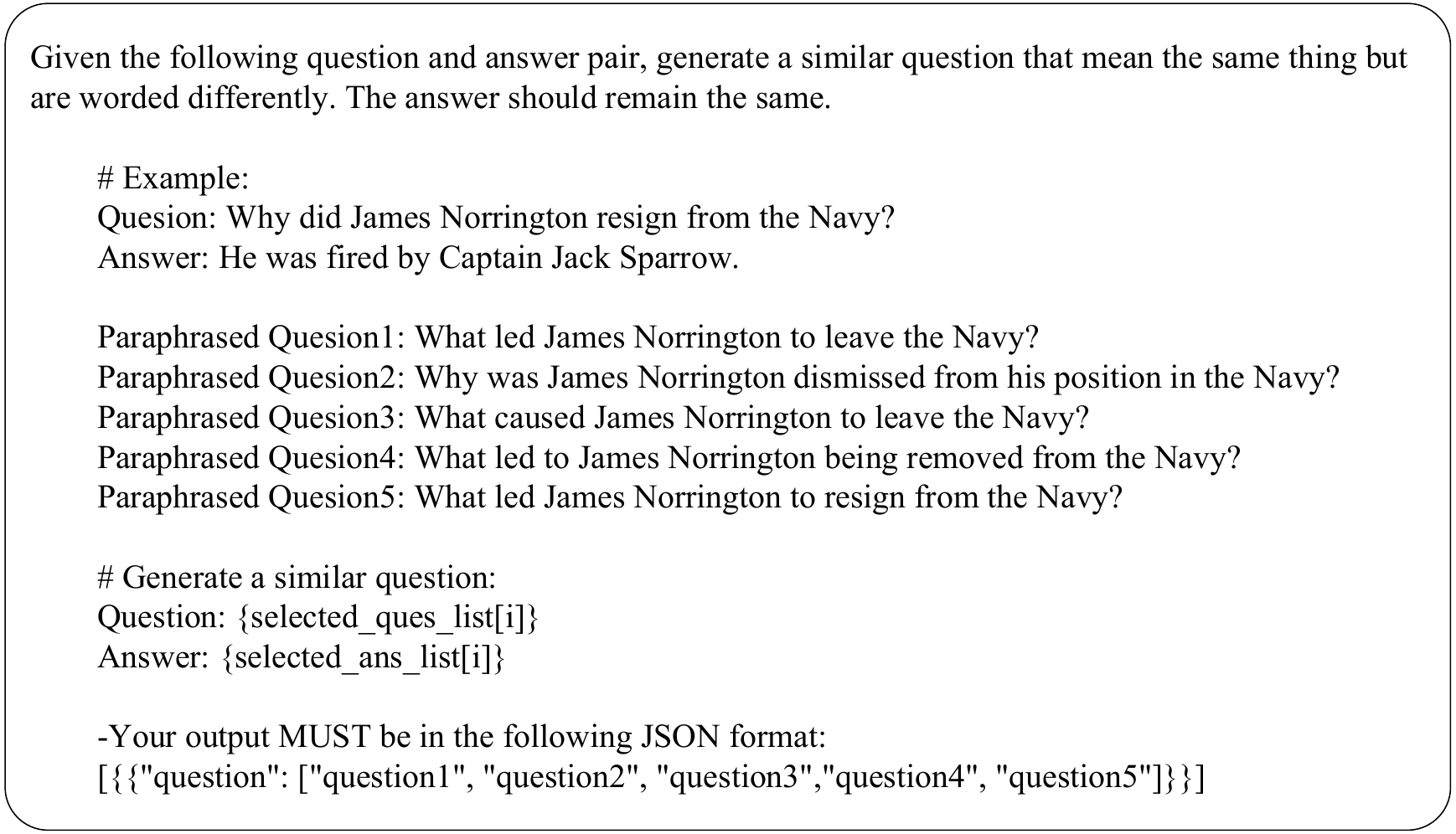}  
    \caption{Prompt for paraphrased question generation}
    \label{fig:prompt_para_q}
\end{figure}

\subsection{Prompt for Seed Topics Generation}
\label{subsec:prompt_seed_topics}
\begin{figure}[!ht]
    \centering
    \includegraphics[width=\textwidth]{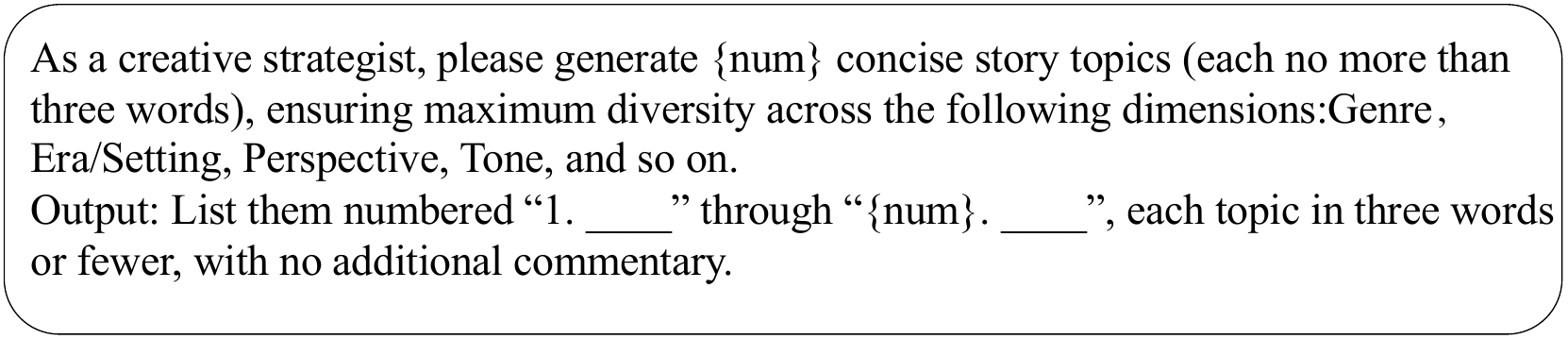}  
    \label{fig:prompt_topic}
\end{figure}

\subsection{Prompt for Topic-guided Generation}
\label{subsec:prompt_topic}
\begin{figure}[!ht]
    \centering
    \includegraphics[width=\textwidth]{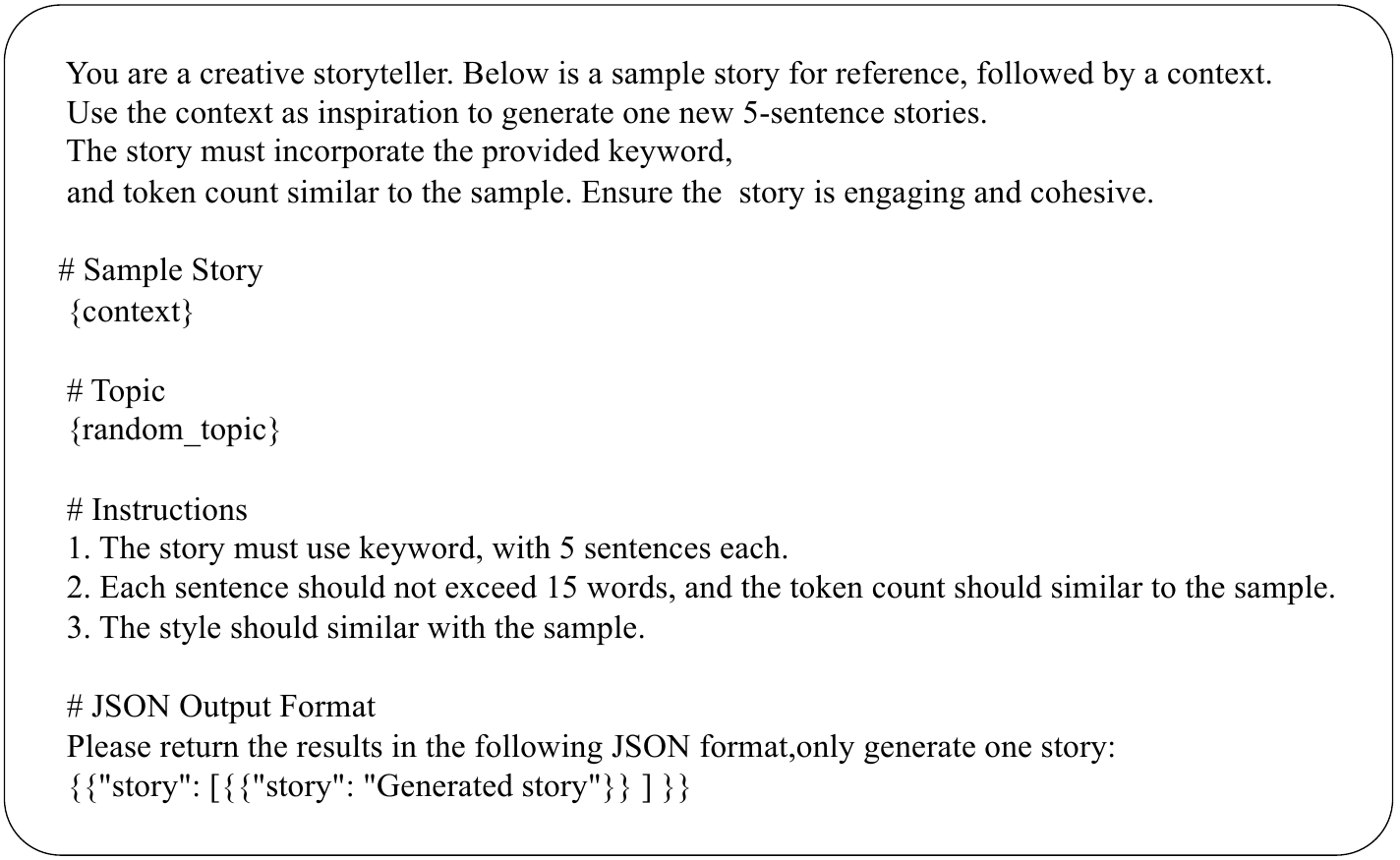}  
    \label{fig:prompt_roc}
\end{figure}

\subsection{Topics.}
\label{subsec:topics}
We present some topics generated by GPT-4o below:

"Adventure", "Love", "Friendship", "Betrayal", "Horror", "Dreams", "Self-discovery", "War", "Sacrifice", "Growth", 
"Courage", "Revenge", "Redemption", "Humanity", "Choice", "Ethics", "Morality", "Family", "Loss", "Hope", 
"Politics", "Race", "Future", "Technology", "Supernatural", "Mystery", "Fantasy", "Magic", "Humor", "Happiness", 
"Pain", "Escapism", "Responsibility", "Fate", "Inner world", "Society", "Culture", "Emotion", "Revival", "Apocalypse", 
"Time travel", "Crime", "Psychology", "Paranormal", "Miracles", "Darkness", "Light", "Illusion", "Life and death", "Truth",
"Grudges", "Prophecy", "Legend", "Myth", "Memory", "Lost", "Freedom", "Taboos", "Dystopia", "Utopia",
"Powerful", "Weak", "Fleeing", "Massacre", "Pursuit", "Omen", "Conspiracy", "Evil deeds", "Justice", "Rebellion", "Forbidden love", "Madness", "Survival", "Loneliness", "Mystical island", "Mysterious disappearance", "Demon King", "Fairy", 
"Aliens", "Machines", "Robots", "Digital world", "Ancient civilization", "Ancient ruins", "Ritual", "Trap", "Kinship", "Labyrinth", 
"Gangster", "Espionage", "Business war", "Betrayal", "Impossible mission", "Evolution", "Civil war", "Empire", "Nation", "Rebirth"

\section{Related Work of LLM-based Data Generation}
\label{apd:related_work}
Inspired by the exceptional generative capabilities of LLMs, researchers have proposed numerous strategies to harness LLMs as data generators~\cite{wang2022self,yu2024large}. In the early stages, several studies~\cite{ding2022gpt,yoo2021gpt3mix,dai2023auggpt,gilardi2023chatgpt} were dedicated to answering whether ChatGPT or GPT is capable of data augmentation or data annotation. Follow-up studies have attempted to leverage LLMs to augment~\cite{ding2024data}, annotate~\cite{tan2024large}, and generate data based on seed samples~\cite{li2023synthetic} or from scratch~\cite{ye2022zerogen}. However, most of these methods, such as ZeroGen~\cite{ye2022zerogen}, ProGen~\cite{ye2022progen}, and SuperGen~\cite{meng2022generating}, primarily focus on text classification tasks and rely heavily on prompt design to enhance data generation performance. To further improve performance and extend applications, advanced strategies have been developed. For instance, self-instruct~\cite{wang2022self} first prompts LLMs to generate instructions, which are then used to prompt the generation of data instances. AttrPrompt~\cite{yu2024large} leverages diverse attributed prompts to eliminate biases of generated data while enhancing its diversity. To address the need for specific-domain instances as seed samples, TarGen~\cite{gupta2023targen} generates seed samples for the specific domain through LLMs, followed by data instance generation with self-correction. As LLMs have become powerful tools for enriching training data, many foundational LLMs, such as Llama 3~\cite{dubey2024llama}, phi-4~\cite{abdin2024phi}, and GPT-4~\cite{achiam2023gpt}, incorporate LLM-generated data into the post-training. Strategies like self-revision and instruction reversal are employed to ensure the quality of generated data. In contrast, our work aims to investigate model performance under generated data with varying diversity, involving controlling the diversity of generated data. 

\section{Details of Datasets}
\label{sec:datasets}
We utilize NarrativeQA~\cite{narrativeqa} and ROCStories~\cite{rocstories} datasets as our real-world datasets, corresponding to long context question-answering and story completion tasks, respectively. The dataset statistics are detailed in Table~\ref{tab:data_stat}. 
Due to the large volume of samples in NarrativeQA and ROCStories, we select a portion of the original datasets for our real-world training data. For each long context in NarrativeQA, we randomly sample 30 question-answering pairs for Sections~\ref{subsec:var_diver} and 20 question-answering pairs for Section~\ref{subsec:mix_ratio}. For ROCStories, we extract 30,000 samples from the training set of the original dataset for our training purposes, and 10,000 samples from the test set of the original dataset for our testing purposes. Each sample from ROCStories is further split into story context and completion, with the first four sentences as the story context and the last sentence as the story completion. 

To create LLM-generated or synthetic datasets, we employ GPT-4o as the data generator, setting the temperature to 0.7 and using samples from the aforementioned datasets as seed examples. To construct LLM-generated data with various diversity, we employ data generation pipelines in Section~\ref{subsubsec:data_gen} to construct seven diversity levels of LLM-generated data, named from \textit{Gen-D0} (the lowest diversity) to \textit{Gen-D6}  (the highest diversity). Specifically, we leverage paraphrasing-based augmentation for long context QA dataset generation, where $m_{p}$ is set to 5 and $m_{r}$ is set as $\{30, 25, 20, 15, 10, 5, 0\}$ for \textit{Gen-D0/1/2/3/4/5/6}, respectively. We utilize topic-guided generation for the story completion dataset generation, where $m_{t}$ is set as $\{20, 100, 300, 1k, 3k, 6k, 10k\}$ for \textit{Gen-D0/1/2/3/4/5/6}, respectively. We prompt LLMs to generate 10k topics, from which we randomly sample a subset based on $m_{t}$. We provide part of the topics in Appendix~\ref{subsec:topics}.

\section{Additional Experiments} 
\label{apd:add_exp}

\subsection{Impact of Model Architecture}
\label{subsec:impact_model_arch}

\textbf{Setup.} We investigate the impact of data diversity on model performance across different model architectures. Specifically, we follow the experimental settings in Sections~\ref{subsec:var_diver} and~\ref{subsec:mix_ratio}, and repeat experiments across three different architectures, including OPT-1.3B, Llama 3.2-1B, and GPT2 xl-1.5 B. Notably, we use models with a similar size to avoid the impact of model capacities.

\begin{figure}[!tbp]
    \centering
    \begin{subfigure}
        \centering
        \includegraphics[width=0.45\linewidth]{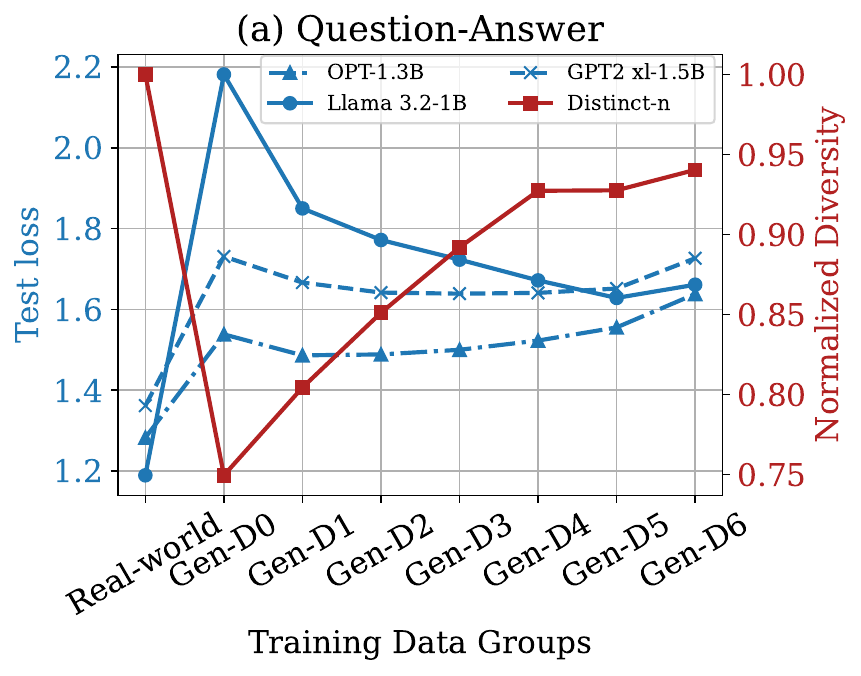}
        \label{fig:arch_divers_qa}
    \end{subfigure}
    \begin{subfigure}
        \centering
        \includegraphics[width=0.45\linewidth]{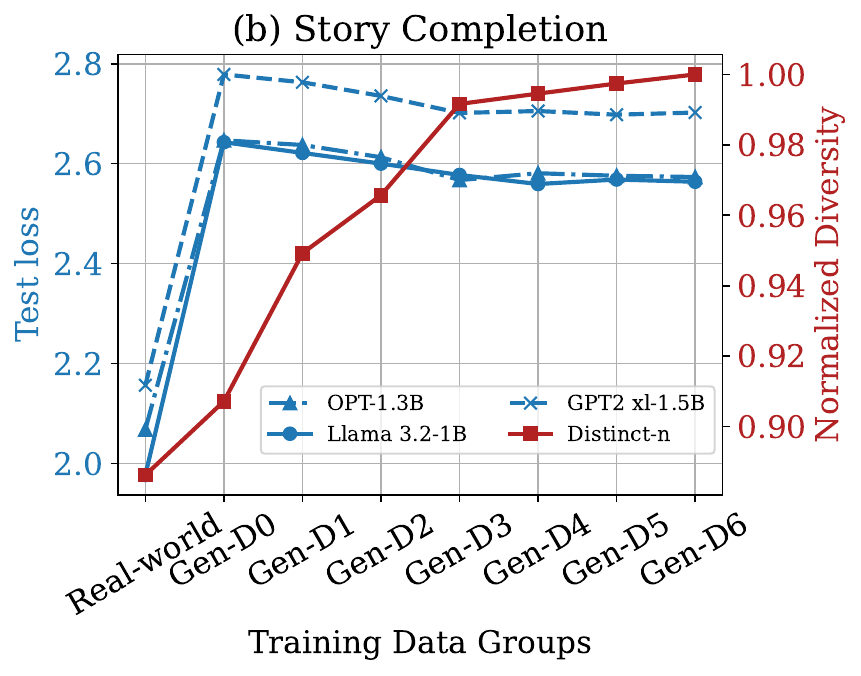}
        \label{fig:arch_divers_story}
    \end{subfigure}
    \caption{The diversity of LLM-generated data and model performance under various model architectures. (a), (b) Experimental results on the question-answering task (story completion task).}
    \label{fig:arch_divers_all}
\end{figure}

\begin{figure}[!tbp]
    \centering
    \begin{subfigure}
        \centering
        \includegraphics[width=0.45\linewidth]{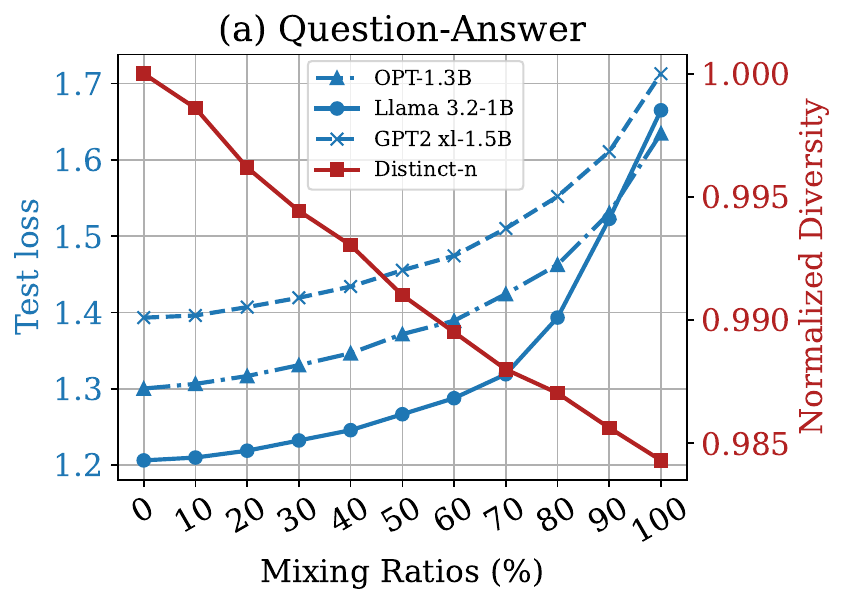}
        \label{fig:arch_mix_qa}
    \end{subfigure}
    \begin{subfigure}
        \centering
        \includegraphics[width=0.45\linewidth]{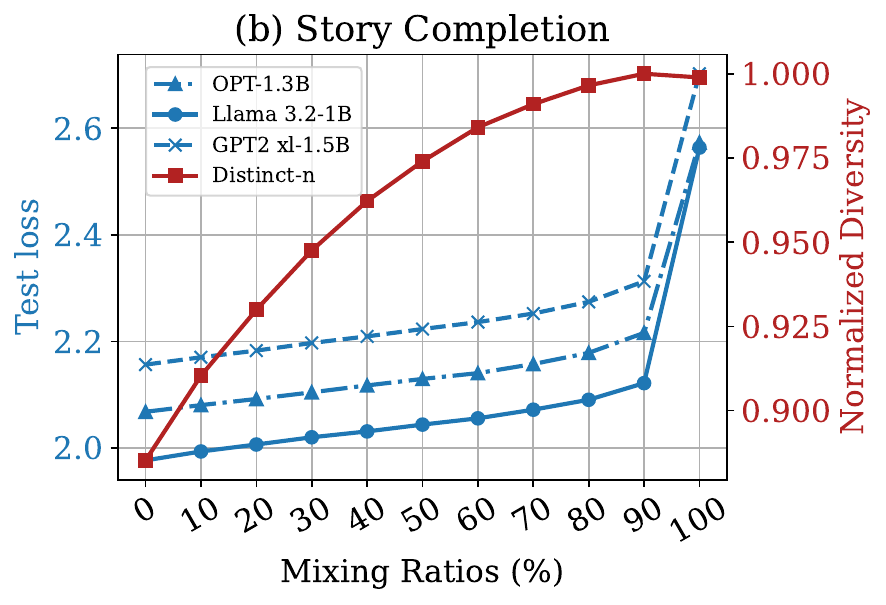}
        \label{fig:arch_mix_story}
    \end{subfigure}
    \caption{Model performance and dataset diversity under different mixing ratios across various architectures. (a), (b) Experimental results on the question-answering task (story completion task).}
    \label{fig:arch_mix_ratio_all}
\end{figure}

\textbf{Results.}
Figures~\ref{fig:arch_divers_all} and~\ref{fig:arch_mix_ratio_all} illustrate the performance of models trained on LLM-generated data with varying diversity and synthetic data with different mixing ratios, respectively. From these two figures, we observe similar experimental results as those of Sections~\ref{subsec:var_diver} and~\ref{subsec:mix_ratio}, providing strong support for the conclusions drawn in these sections. Moreover, the consistent trends observed across different models suggest that the impact of model architecture is negligible.

\subsection{Impact of Model Size}
\label{subsec:impact_model_size}

\begin{figure}[!tb]
    \centering
    \includegraphics[width=0.95\linewidth]{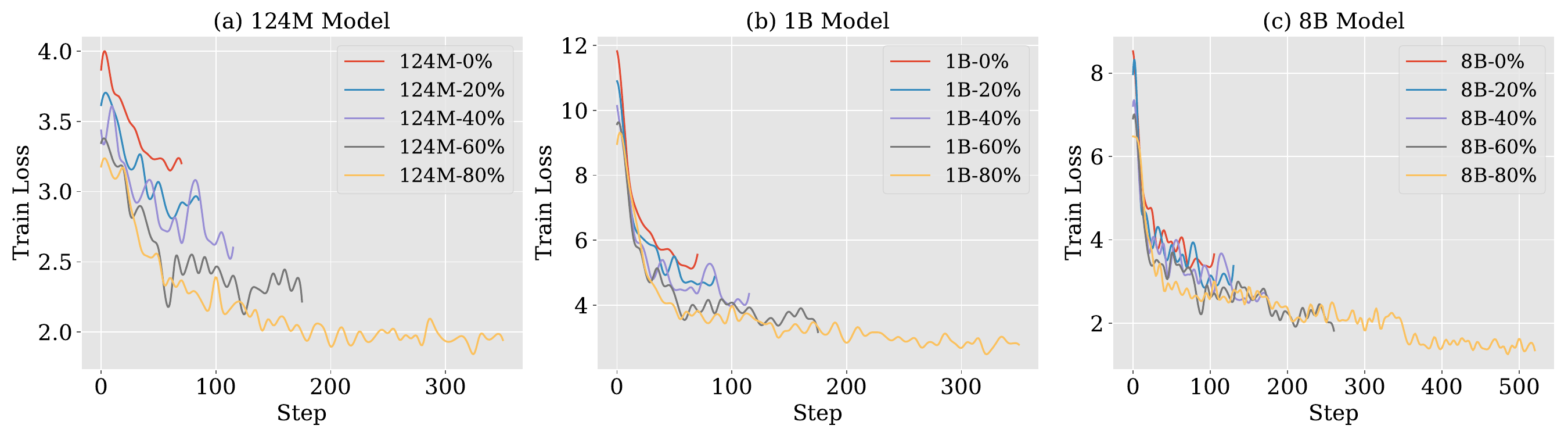}
    \caption{Loss curves corresponding to Gen-D5-sourced models in Figure~\ref{fig:mix_vary_qa}.}
    \label{fig:training_loss}
\end{figure}

\textbf{Setup.} We investigate the impact of data diversity on model performance across various model sizes. Specifically, we further analyze the experimental results of Sections~\ref{subsec:var_diver} and~\ref{subsec:mix_ratio}, involving three model sizes, i.e., 124M, 1B, and 8B.

\textbf{Results.}
Generally, we observe consistent relationships between model performance and the diversity of LLM-generated data across various model sizes. However, as shown in Figure~\ref{fig:mix_vary_qa}, the results differ significantly with different model sizes. Specifically, the performance of the 124M and 8B models exhibits opposite correlations with data diversity: \textit{the 124M model's performance is positively correlated with data diversity, while the 8B model's performance is negatively correlated}. The 1B model is in an intermediate state between the 124M and 8b models. As the mixing ratios increase, the diversity of synthetic data decreases, indicating that the newly added LLM-generated data bears a high similarity to the pre-existing real data. Consequently, the model encountering more similar data within the same epoch effectively equates to an increase in the number of epochs. In other words, as the mixing ratios increase, the number of epochs is implicitly increased during model training. Larger models have stronger data fitting capabilities, so for the 8B model, this implicit increase in epochs leads to overfitting. Conversely, the 124M model, with its weaker fitting ability, benefits from the additional epochs, enhancing data fitting and performance. As shown in Figure~\ref{fig:training_loss}, we observe a sudden drop in the training loss for the 8B-60\% and 8B-80\% models, indicating a strong fit to the training data after each epoch. This pattern suggests that the 8B models are experiencing overfitting.



\end{document}